\documentclass{article}

\usepackage{arxiv}

\usepackage[utf8]{inputenc} 
\usepackage[T1]{fontenc}    
\usepackage{hyperref}       
\usepackage{url}            
\usepackage{booktabs}       
\usepackage{nicefrac}       
\usepackage{microtype}      
\usepackage{lipsum}
\usepackage{colortbl}  
\usepackage{xcolor}
\usepackage{algorithmicx}
\usepackage{multicol}
\usepackage{multirow}
\usepackage{algorithm}
\usepackage{caption}
\usepackage{amsmath,amssymb,amsfonts}
\usepackage{amsthm} 
\usepackage{graphicx}
\usepackage{algpseudocode}
\graphicspath{ {./images/} }

\title{UISearch: Graph-Based Embeddings for Multimodal Enterprise UI Screenshots Retrieval}

\author{
 Maroun Ayli \\
  Center For Computer Science\\
  Saint Joseph University of Beirut\\
  Beirut Lebanon \\
  \texttt{maroun.ayli1@usj.edu.lb} \\
   \And
 Youssef Bakouny \\
  Center For Computer Science\\
  Saint Joseph University of Beirut\\
  Beirut Lebanon \\
  \texttt{youssef.bakouny@usj.edu.lb} \\
  \And
   Tushar Sharma \\
  Faculty of Computer Science\\
  Dalhousie University\\
  Halifax, Canada \\
  \texttt{tushar@dal.ca} \\
  \And
  Nader Jalloul \\
  Murex\\
  Paris, France \\
  \texttt{njalloul@murex.com} \\
  \And
  Hani Seifeddine \\
  Murex\\
  Paris, France \\
  \texttt{hseifeddine@murex.com} \\
  \And
  Rima Kilany \\
  Center For Computer Science\\
  Saint Joseph University of Beirut\\
  Beirut Lebanon \\
  \texttt{rima.kilany@usj.edu.lb} \\
}

\begin{document}

\maketitle

\begin{abstract}
Enterprise software companies maintain thousands of user interface screens across products and versions, creating critical challenges for design consistency, pattern discovery, and compliance check. Existing approaches rely on visual similarity or text semantics, lacking explicit modeling of structural properties fundamental to user interface (UI) composition. We present a novel graph-based representation that converts UI screenshots into attributed graphs encoding hierarchical relationships and spatial arrangements, potentially generalizable to document layouts, architectural diagrams, and other structured visual domains. A contrastive graph autoencoder learns embeddings preserving multi-level similarity across visual, structural, and semantic properties. The comprehensive analysis demonstrates that our structural embeddings achieve better discriminative power than state-of-the-art Vision Encoders, representing a fundamental advance in the expressiveness of the UI representation. We implement this representation in UISearch, a multi-modal search framework that combines structural embeddings with semantic search through a composable query language. On 20,396 financial software UIs, UISearch achieves 0.92 Top-5 accuracy with 47.5ms median latency (P95: 124ms), scaling to 20,000+ screens. The hybrid indexing architecture enables complex queries and supports fine-grained UI distinction impossible with vision-only approaches.
\end{abstract}
    
\section{Introduction}
    
    Enterprise software systems contain tens of thousands of user interface screens across applications, versions, and platforms. Design teams face unprecedented challenges in ensuring consistency across products, discovering reusable patterns, identifying compliance violations, and maintaining design system adherence. Current approaches rely on manual inspection or simple visual similarity tools, failing to capture the semantic and structural properties essential for effective UI management.
    
    Consider searching for ``payment screens with forms containing 3-5 input fields but excluding data tables.'' Existing tools can perform basic visual matching or keyword search, but cannot analyze structural composition or combine these modalities. Effective UI retrieval requires understanding across multiple dimensions: (1) visual appearance, the pixel-level similarity of screens; (2) structural composition, the arrangement and relationships between UI elements; (3) semantic intent, the functional purpose of the screen; and (4) metadata constraints, specific requirements about UI composition.
    
    Recent advances have made progress on individual aspects. Screen2Vec~\cite{screen2vec} introduced semantic embeddings but lacks structural awareness. LayoutGMN~\cite{layoutgmn} models layout similarity but ignores functional intent. Document embedding approaches handle visual similarity but cannot support structured queries. No existing system provides comprehensive multimodal search for enterprise UI collections.
    
    We present UISearch, a multimodal search engine designed for enterprise UI management. Unlike traditional approaches, UISearch converts screenshots into graph representations that capture element relationships and spatial arrangements. A contrastive graph autoencoder learns embeddings preserving multi-level similarity, while a hybrid indexing architecture enables subsecond queries combining visual, structural, semantic, and metadata constraints.
    
    Our key contributions are:
    \begin{itemize}
        \item \textbf{Graph-Based UI Representation:} We develop a comprehensive graph representation capturing both hierarchical structure and spatial relationships of UI elements, feeding into a contrastive graph autoencoder that achieves better discriminative power than state of the art VLM embeddings.
        \item \textbf{Multimodal Query Framework:} We introduce the first UI search system that supports queries combining visual similarity, structural patterns, semantic intent, and metadata constraints through a composable query language.
        \item \textbf{Scalable Indexing Architecture:} Our hybrid approach combines FAISS vector indices with metadata filtering, achieving subsecond query latency at scale.
        \item \textbf{Enterprise Deployment:} We demonstrate real-world effectiveness on 20,000+ enterprise financial UIs with comprehensive evaluation including performance benchmarks and comparisons with existing approaches.
    \end{itemize}
    
    \section{Related Work}
    
    \textbf{UI Datasets and Embeddings.} The Rico dataset~\cite{ricodataset} provided 72,000 Android UI screens, establishing the foundation for computational UI analysis. Rico exposed visual, textual, structural, and interactive design properties, demonstrating UI layout similarity using autoencoders. Screen2Vec~\cite{screen2vec} advanced semantic understanding by introducing comprehensive embeddings combining textual content, visual layout, and app context through self-supervised learning, using user interaction traces as context instead of manual labels. Recent improvements include CLAY~\cite{clay}, which applied deep learning for layout denoising at scale with 59,555 human-annotated screens achieving 82.7\% F1 score for invalid object detection, and MUD~\cite{mud}, providing 18,000 human-annotated UIs from 3,300 apps using large language models for automated exploration. While these approaches capture semantic properties, they lack explicit structural modeling essential for enterprise search.
    
    \textbf{Graph-Based UI Representations.} LayoutGMN~\cite{layoutgmn} established graph-based structural similarity matching using Graph Matching Networks to predict similarity between 2D layouts including UI designs. The attention-based GMN in triplet network setting with IoU-based weak supervision provided the first deep model offering both metric learning of structural layout similarity and element-level matching. Graph4GUI~\cite{graph4gui2024} exploited graph neural networks that capture both semantic properties of UI elements and their visuo-spatial constraints, with embeddings excelling at tasks like GUI autocompletion where it predicted placements of missing elements with layouts preferred by designers. Screen Parsing~\cite{wu2021screenparsing} used computer vision to detect UI elements and their hierarchies from screenshots, effectively reverse-engineering the view hierarchy to enable applications such as UI similarity search based on structural content. These graph-oriented approaches complement pixel/text embeddings by focusing on relational layout patterns, particularly useful for detecting design analogies between screens with different content but similar arrangements.
    
    \textbf{Multimodal Retrieval and Fusion.} Contrastive learning emerged as the dominant paradigm for aligning different modalities in retrieval systems. Research on understanding the modality gap~\cite{mindthegap} showed that different modalities naturally embed in separate regions, and manipulating this separation can improve zero-shot performance. DQU-CIR shifted multimodal fusion from feature-level to raw-data level, preventing fused features from deviating from original embedding spaces~\cite{wen2024simple}. Set-based embeddings addressed inherent ambiguity in cross-modal retrieval by encoding samples into multiple embedding vectors capturing different semantics~\cite{multimodalfusion}, proving particularly relevant for UI search where interfaces can be interpreted from multiple perspectives.
    
    \textbf{Vision-Language Models.} Recent VLMs have dramatically improved UI understanding. ScreenAI~\cite{baechler2024screenai} achieved breakthrough performance on UI and infographics understanding using specialized vision-language architecture with flexible patching at 5 billion parameters, establishing new state-of-the-art results on Multipage DocVQA, WebSRC, MoTIF, and Widget Captioning. Spotlight~\cite{li2023spotlight} eliminated dependency on view hierarchy information using vision-only approaches, with its Focus Region Extractor concentrating on specific UI regions while maintaining screen context. Widget captioning research~\cite{li2020widget} established foundational work in automatically generating language descriptions for UI elements. UIClip~\cite{uiclip2024} adapted CLIP for UI design assessment by jointly embedding screenshots and metadata descriptions, enabling both quality ranking and similarity search across large screen repositories. However, our analysis reveals that general-purpose VLMs like CLIP exhibit poor discriminative power for enterprise UI search.
    
    \textbf{Object Detection for UI Elements.} Mobile UI element detection adapted computer vision for interface-specific challenges. APT (Adaptively Prompt Tuning)~\cite{gu2023mobile} leveraged OCR information alongside visual features using CLIP-based detectors. The UIED toolkit~\cite{xie2020uied} integrated multiple detection methods with interactive dashboards, achieving state-of-the-art F1 scores while exporting results to design tools. Comprehensive empirical studies~\cite{chen2020object} showed that combining traditional CV methods with deep learning outperformed pure deep learning approaches for GUI-specific detection tasks. YOLOv5-MGC~\cite{2022yolov5gui} achieved 89.8\% mAP using K-means++ anchor generation, CBAM attention mechanisms, and microscale detection layers for tiny elements.
    
    \textbf{Scalable Similarity Search.} FAISS~\cite{douze2024faiss} provides optimized k-means and product quantization on GPUs, achieving 8$\times$ speedups over prior methods and building indexes of 1B+ vectors searchable in sub-second latency. Graph-based approximate nearest neighbor methods dominated high-accuracy scenarios. HNSW~\cite{malkov2018efficient} achieved logarithmic complexity scaling through multi-layer proximity graph structures. These ANN indexes have become backbone infrastructure for scalable UI search, allowing enterprise-scale screen collections to be indexed by learned embeddings and queried interactively.
    
    Our work addresses the critical gap between these individual advances and comprehensive enterprise UI search systems by integrating multimodal similarity computation, graph-based structural analysis, vision-language understanding, and scalable indexing in a unified query framework.

\section{Methods}
We formalize an end-to-end pipeline that lifts raw screens into a typed graph space, learns structure-aware embeddings, and executes hybrid retrieval under enterprise constraints.
\subsection{System Overview and Design Principles}

UISearch employs a mathematically principled architecture designed to handle multimodal queries over large-scale UI collections. Let $\mathcal{U} = \{u_1, u_2, ..., u_n\}$ denote a collection of $n$ UI screens, where each $u_i$ consists of a screenshot image $I_i \in \mathbb{R}^{H \times W \times 3}$ and its corresponding structural representation.

We define UISearch as a tuple $\mathcal{S} = (\mathcal{U}, \mathcal{G}, \mathcal{E}, \mathcal{I}, \mathcal{Q}, \mathcal{R})$ where:
\begin{itemize}
    \item $\mathcal{U}$: Universe of UI screens
    \item $\mathcal{G}: \mathcal{U} \rightarrow G$: Graph construction function
    \item $\mathcal{E}: G \rightarrow \mathbb{R}^d$: Embedding function
    \item $\mathcal{I}$: Indexing structure over embeddings
    \item $\mathcal{Q}$: Query language with formal syntax
    \item $\mathcal{R}: \mathcal{Q} \times \mathcal{U} \rightarrow [0,1]$: Ranking function
\end{itemize}

\begin{figure}[htbp]
\centering
\includegraphics[width=0.6\columnwidth]{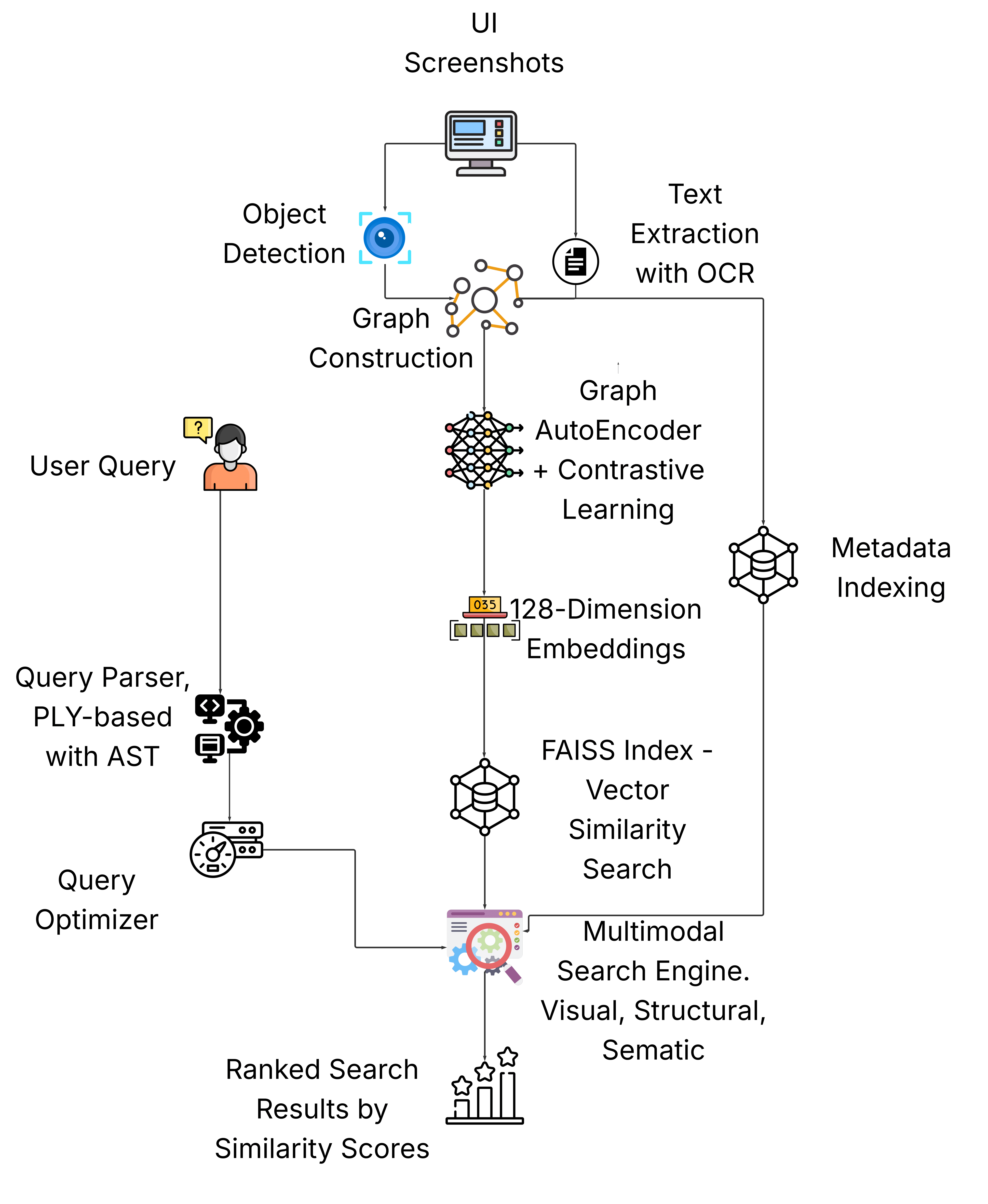}
\caption{UISearch system architecture showing the complete pipeline from UI screenshots through graph construction, contrastive learning, and hybrid indexing to ranked search results.}
\label{fig:architecture}
\end{figure}

Our system adheres to three fundamental principles: \textbf{P1 (Completeness)} - the system must find all relevant results for any valid query; \textbf{P2 (Efficiency)} - query processing must be fast enough for interactive use with $T(q) = O(\log n + k)$ where $k$ is result size; \textbf{P3 (Compositionality)} - complex queries should decompose into primitive operations.  Figure~\ref{fig:architecture} illustrates the complete system pipeline.

\subsection{Graph-Based UI Representation}

Traditional approaches treat interfaces as flat images or simple element lists, losing critical structural information. Our graph representation preserves spatial relationships, hierarchical containment, and interaction patterns that define interface structure.

For each UI screen $u_i$, we construct a directed graph $G_i = (V_i, E_i, X_i, A_i)$ where $V_i$ represents UI elements, $E_i$ encodes spatial relationships, $X_i \in \mathbb{R}^{|V_i| \times f}$ contains node features with $f=16$ features per element, and $A_i \in \mathbb{R}^{|V_i| \times |V_i|}$ is the weighted adjacency matrix.

\textbf{Element Detection.} We employ a specialized YOLO model ~\cite{yolov8} trained specifically on UI screenshots to detect and classify 15 element types: Label, Button, Dropdown, Table, Menu Item, Radio Button, Icon, Links, CheckBox, OptionsButton, WindowName, Icon Button, Text Box, DatePicker, and Window. This classification balances granularity with practical utility while avoiding over-specification.

\textbf{Feature Engineering.} For each UI element $v_j \in V_i$, we extract a comprehensive feature vector $x_j \in \mathbb{R}^{16}$ combining spatial, type, and interaction properties. Spatial features capture normalized position, size, area, and aspect ratio relative to screen dimensions. Type features use one-hot encoding for the 9 most common element categories. The interaction features distinguish actionable from informational elements.

\textbf{Edge Construction.} The edges of our user interface graphs encode meaningful spatial relationships between interface elements. Simply connecting every element to every other element would create an overly dense graph that obscures important patterns. Instead, we selectively create edges based on spatial proximity and functional relationships.
Edges are created on the basis of both distance and overlap criteria. For element pairs $(v_j, v_k)$:
\begin{equation}
(v_j, v_k) \in E_i \Leftrightarrow d(v_j, v_k) < \theta_d \vee \text{IoU}(v_j, v_k) > \theta_{\text{IoU}}
\end{equation}
where $d(v_j, v_k)$ is Euclidean distance between centers, $\theta_d = 0.25\sqrt{W^2 + H^2}$ adapts to screen size, and $\theta_{\text{IoU}} = 0.1$ captures meaningful overlaps. Edge weights combine distance similarity, type similarity, and IoU with weights $\alpha = 0.6, \beta = 0.3, \gamma = 0.1$ prioritizing spatial proximity.

Algorithm \ref{alg:graph_construction} details the construction process from images to graphs.

\begin{algorithm}[h]
\caption{UI Graph Construction Pipeline}
\label{alg:graph_construction}
\begin{algorithmic}[1]
\State \textbf{Input:} UI screenshot $I$, detection threshold $\tau_{\text{conf}}$, distance threshold $\tau_d$
\State \textbf{Output:} Graph $G = (V, E, \mathbf{X}, \mathbf{A})$
\State $\mathcal{D} \leftarrow \text{YOLO}_{\text{UI}}(I, \tau_{\text{conf}})$ \Comment{Detect UI elements}
\State $V \leftarrow \emptyset$, $E \leftarrow \emptyset$, $\mathbf{X} \leftarrow []$
\For{each detection $d_i \in \mathcal{D}$}
   \State $\mathbf{b}_i \leftarrow \text{BoundingBox}(d_i)$
   \If{$\text{Type}(d_i) \in \{\text{label}, \text{textbox}, \text{window}\}$}
       \State $t_i \leftarrow \text{OCR}(\text{Crop}(I, \mathbf{b}_i))$
   \Else
       \State $t_i \leftarrow \text{Type}(d_i)$
   \EndIf
   \State $v_i \leftarrow \text{CreateNode}(d_i, \mathbf{b}_i, t_i)$
   \State $\mathbf{x}_i \leftarrow \text{ExtractFeatures}(v_i, I)$
   \State $V \leftarrow V \cup \{v_i\}$, $\mathbf{X} \leftarrow \mathbf{X} \oplus \mathbf{x}_i^T$
\EndFor
\State $E, \mathbf{A} \leftarrow \text{ConstructEdges}(V, \tau_d)$
\State \textbf{return} $G = (V, E, \mathbf{X}, \mathbf{A})$
\end{algorithmic}
\end{algorithm}

\subsection{Contrastive Graph Autoencoder}

Our encoder $f_\theta: \mathcal{G} \rightarrow \mathbb{R}^{2d}$ combines Graph Attention Networks with Graph Convolutional Networks to capture both local attention patterns and global structure.

\textbf{Encoder Architecture.} The architecture employs two GAT layers with 4 attention heads each, followed by layer normalization, ReLU activation, and dropout ($p = 0.1$), then a final GCN layer for global aggregation. Standard GAT attention mechanisms compute attention coefficients using LeakyReLU activation.

\textbf{Multi-Scale Pooling.} We obtain graph-level representations by combining mean pooling (capturing average characteristics), max pooling (preserving salient features), and attention-weighted pooling (focusing on important elements). The final representation concatenates mean and max pooling:
\begin{equation}
\mathbf{g} = [\mathbf{g}_{\text{mean}} \oplus \mathbf{g}_{\text{max}}] \in \mathbb{R}^{2d}
\end{equation}

\textbf{Contrastive Learning Framework.} For a batch $\mathcal{B} = \{G_1, \ldots, G_B\}$, we compute pairwise similarities $\mathbf{S} \in \mathbb{R}^{B \times B}$ using our multi-level similarity function. We project to normalized space and optimize:
\begin{equation}
\mathcal{L}_{\text{contrastive}} = -\frac{1}{|\mathcal{P}|} \sum_{(i,j) \in \mathcal{P}} \log \frac{\exp(\mathbf{p}_i^T \mathbf{p}_j / \tau)}{\sum_{k \neq i} \exp(\mathbf{p}_i^T \mathbf{p}_k / \tau)}
\end{equation}
where $\mathcal{P} = \{(i,j) : S_{ij} > \theta_{\text{pos}}\}$ are positive pairs and $\tau=0.1$ is temperature.

\textbf{Multi-Task Learning.} We jointly optimize contrastive learning with intent classification and graph reconstruction:
\begin{equation}
\mathcal{L}_{\text{total}} = \mathcal{L}_{\text{contrastive}} + \lambda_{\text{intent}}\mathcal{L}_{\text{intent}} + \lambda_{\text{recon}}\mathcal{L}_{\text{reconstruction}}
\end{equation}
Intent classification predicts functional categories (login, checkout, dashboard, etc.) using a softmax classifier over graph embeddings. Graph reconstruction uses binary cross-entropy to reconstruct the adjacency matrix from node embeddings.

\textbf{Training Strategy.} We employ adaptive positive/negative mining with curriculum learning where thresholds evolve during training, progressively refining positive and negative pair definitions.

\subsection{Multi-Level Similarity Metrics}

Our comprehensive similarity function combines four intuitive measures weighted equally:
\begin{equation}
S(G_i, G_j) = \sum_{k=1}^{4} w_k \cdot S_k(G_i, G_j)
\end{equation}
Type similarity uses Jaccard overlap of element types, size similarity compares interface complexity via element count ratios, density similarity captures layout connectivity differences, and interactive similarity compares the proportion of actionable elements.

\subsection{Hybrid Indexing and Query Processing}

Our indexing system maintains multiple specialized indices and routes queries based on their characteristics.

\textbf{Dual FAISS Implementation.} We maintain separate indices for semantic and structural similarity. Semantic embeddings from sentence transformers~\cite{reimers2019sentencebert} are L2-normalized and indexed using IndexFlatIP for cosine similarity. Structural embeddings support cosine, Euclidean, and inner product metrics with query-time selection.

\textbf{Metadata Indexing.} For metadata filtering, we construct inverted indices for exact count matching, sorted indices for range queries, and boolean indices for containment queries. These enable efficient filtering by element counts, types, and presence.

\textbf{Query Optimization.} Query execution follows cost-based optimization selecting between FAISS-only (pure similarity), metadata-only (pure filtering), metadata-first (high selectivity constraints), or FAISS-first (default) strategies based on query characteristics and estimated costs.

\textbf{Multimodal Score Fusion.} When queries combine multiple similarity types, we compute dynamic weights:
\begin{equation}
\rho_q(u) = \sum_{m \in \mathcal{M}} \lambda_m \cdot s_m(q, u)
\end{equation}
where $\mathcal{M} = \{\text{visual}, \text{structural}, \text{intent}, \text{metadata}\}$. Weights are normalized based on active query modalities, ensuring that queries using only structural similarity give full weight to that component.

Our proposed architecture (Table~\ref{tab:model_summary}) maintains a compact footprint of just 318,688 trainable parameters (~1.22 MB), enabling efficient training and deployment while preserving strong representation learning capabilities.
\begin{table}[t]
\centering
\caption{Model Architecture Summary}
\label{tab:model_summary}
\begin{tabular}{@{}lrc@{}}
\toprule
\textbf{Layer/Component} & \textbf{Configuration} & \textbf{Params} \\
\midrule
GAT Layer 1 & $16 \xrightarrow{\text{4 heads}} 512$ & 9.7K \\
GAT Layer 2 & $512 \xrightarrow{\text{4 heads}} 512$ & 263.7K \\
GCN Layer 3 & $512 \rightarrow 64$ & 32.8K \\
Projection Head & $64 \rightarrow 128 \rightarrow 32$ & 12.4K \\
\midrule
\textbf{Total} & & \textbf{318.7K} \\
\midrule
Dropout Rate & 0.1 & -- \\
Learning Rate & $10^{-3}$ (AdamW) & -- \\
Contrastive Temp. & $\tau = 0.1$ & -- \\
\bottomrule
\end{tabular}
\end{table}
For 20,000 UIs with 128-dimensional embeddings:
\begin{equation}
\text{Memory}_{\text{total}} = 20{,}000 \times 128 \times 4 \times 2 \approx 20.5 \text{ MB}
\end{equation}    
This modest memory footprint makes the system deployable on standard hardware, even for large UI collections.

\subsection{Complexity Analysis}

We analyze the theoretical complexity of UISearch for processing a query $q$ returning $k$ results from a collection of $n$ UI screens with $d$-dimensional embeddings and $|T|$ element types. Our hybrid architecture achieves sub-linear query time through approximate nearest neighbor search while maintaining practical space requirements.

Table~\ref{tab:complexity} summarizes the asymptotic complexity bounds for all major operations. The system achieves $O(\sqrt{n} + k)$ query time using IVF quantization while total memory requirements are $O(n \cdot (d + |T|))$, approximately 20.5 MB for 20K UIs with $d = 128$ and $|T| = 15$.

\begin{figure*}[htbp]
\centering
\includegraphics[width=\linewidth]{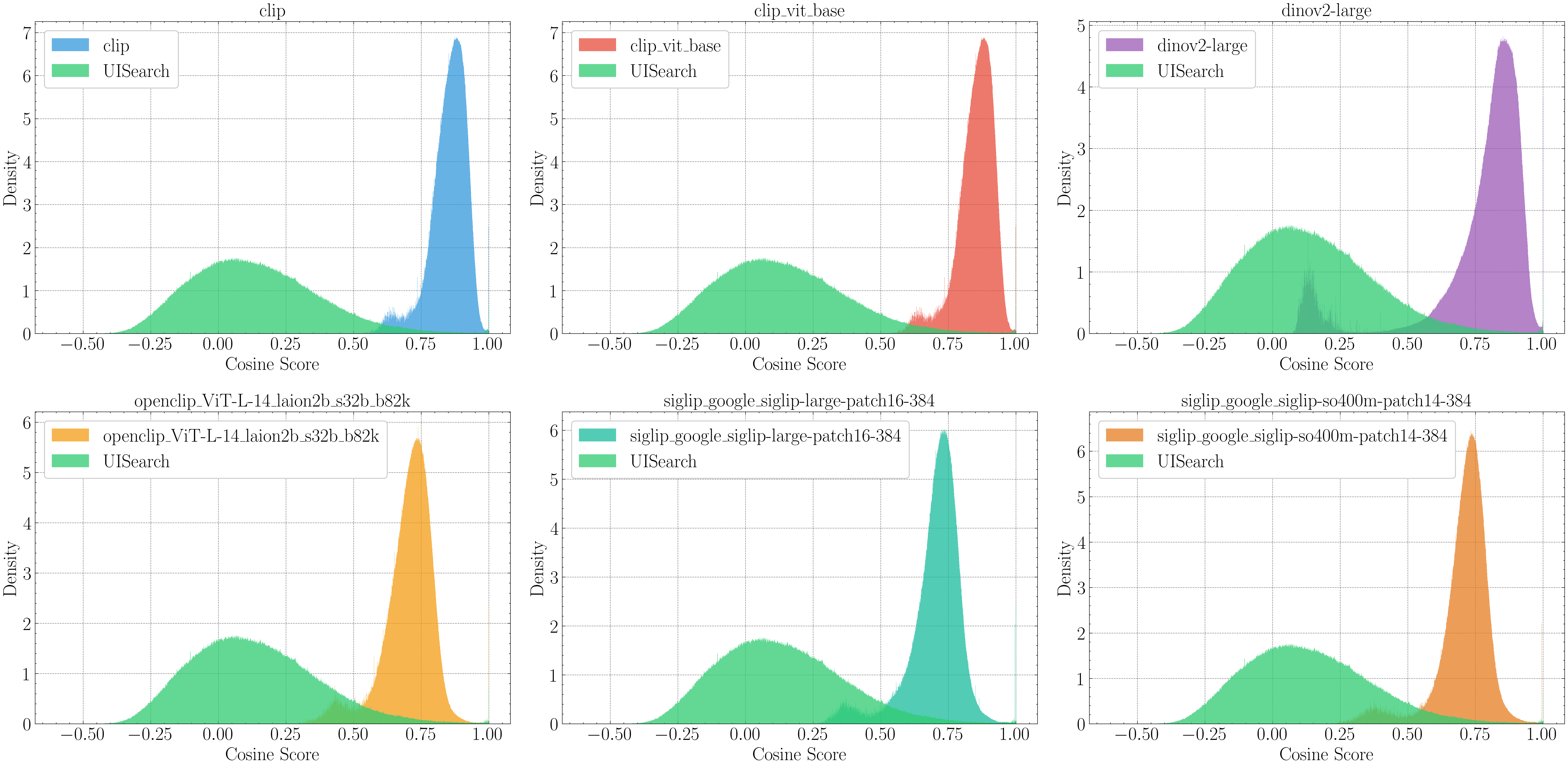}
\caption{UISearch structural embeddings (green) maintain superior image discrimination with broad cosine similarity distributions (0.0-0.4), while vision-language models (CLIP variants, SigLIP, DINOv2) exhibit representation collapse with narrow peaks near 1.0.}
\label{fig:embedding_comparison}
\end{figure*}

\begin{table*}[t]
\centering
\caption{Asymptotic complexity analysis for UISearch query processing ($n$ UI screens, $k$ results, $d$-dimensional embeddings, $|T|$ element types).}
\label{tab:complexity}
\small
\begin{tabular}{@{}p{3.5cm}p{3.5cm}p{7.5cm}@{}}
\toprule
\textbf{Component} & \textbf{Complexity} & \textbf{Description} \\
\midrule
\multicolumn{3}{l}{\cellcolor{gray!20}\textbf{Time Complexity}} \\
Metadata Filtering & $O(|\sigma_{\text{meta}}|)$ & Depends on selectivity of metadata predicates; typically $|\sigma_{\text{meta}}| \ll n$ \\
Vector Search (Approximate) & $O(\sqrt{n} + k)$ & IVF with product quantization; navigates $\sqrt{n}$ clusters then retrieves $k$ results \\
Vector Search (Exact) & $O(n \cdot d)$ & Exhaustive linear scan over all embeddings; fallback for high-precision queries \\
Result Ranking & $O(k \log k)$ & Merge and sort top-$k$ candidates by composite similarity score \\
Hybrid Query (Optimal) & $O(\log n + \sqrt{n} + k)$ & Metadata-first execution with approximate vector search \\
\midrule
\multicolumn{3}{l}{\cellcolor{gray!20}\textbf{Space Complexity}} \\
Dense Embeddings & $O(n \cdot d)$ & Raw 32-bit float vectors for exact computation \\
FAISS Index Structure & $O(n \cdot d/4)$ & Product quantization reduces storage by 4$\times$ with minimal accuracy loss \\
Metadata Indices & $O(n \cdot |T|)$ & Inverted and sorted indices for each of $|T|$ element types \\
\textbf{Total Memory} & $\mathbf{O(n \cdot (d + |T|))}$ & \textbf{Practical: 20.5 MB for 20K UIs with $d{=}128$, $|T|{=}15$} \\
\bottomrule
\end{tabular}
\end{table*}

    \section{Experimental Evaluation}
We benchmark UISearch on production-scale corpora with targeted query suites, strong baselines, and latency-scaling analyses to quantify accuracy and efficiency under realistic load.
    \subsection{Experimental Setup}
    
    \textbf{Dataset.} We evaluate on 20,396 annotated financial software UI screenshots processed via YOLO-based element detection, producing structural graphs and metadata for 15 element types. Test sets contain semantic queries ($|\mathcal{Q}_s|=20$), structural queries ($|\mathcal{Q}_t|=10$), metadata queries ($|\mathcal{Q}_m|=30$), complex multimodal queries ($|\mathcal{Q}_c|=25$), plus ordering, intent, edge-case, and real-world queries for robustness.
    
    \textbf{Metrics.} We report Top-K accuracy and Precision@K for $K \in \{1,3,5,10\}$, Mean Reciprocal Rank (MRR), and latency distributions (P50/P90/P95/P99). We employ vision-language models as automated judges, providing query match decisions, element counts, reasoning, and confidence scores.
    
    \subsection{Retrieval Quality}

    Performance remains consistent across semantic (0.92), structural (0.95), metadata (0.96), and complex multimodal queries (0.90), demonstrating effective integration of multiple modalities.
    
    \subsection{VLM-Based Automated Judging}
    
    We use vision-language models as automated judges, providing the query, retrieved UI screenshot, and (for structural queries) a reference UI image. Each VLM returns query match (true/false), element counts, reasoning, and confidence (0-1).
    
    Table~\ref{tab:vlm_judging} reports per-model judging statistics across 12 models including Qwen2.5-VL, SmolVLM, LLaVA, nvidia.cosmos-reason1-7b, and others. Average acceptance rate is 89.2\% with overall confidence of 0.88, demonstrating strong agreement across diverse VLM architectures.
    
    \begin{table}[t]
    \centering
    \caption{VLM judging performance (selected models).}
    \label{tab:vlm_judging}
    \small
    \begin{tabular}{@{}lccc@{}}
    \toprule
    Model & Time (s) & Accept \% & Confidence \\
    \midrule
    Qwen2.5-VL & 2.1 & 91.2 & 0.90 \\
    SmolVLM & 1.7 & 89.5 & 0.87 \\
    nvidia.cosmos & 2.3 & 92.4 & 0.91 \\
    DeepSeek-VL2 & 1.9 & 90.5 & 0.90 \\
    PaliGemma-2 & 2.3 & 89.0 & 0.88 \\
    gpt-oss-20b & 2.8 & 87.5 & 0.85 \\
    \midrule
    Average (12) & 2.2 & 89.2 & 0.88 \\
    \bottomrule
    \end{tabular}
    \end{table}

    \subsection{Embedding Quality and Discriminative Power}
A fundamental requirement for effective similarity search is that embeddings must meaningfully discriminate between different items. When embeddings fail to distinguish between dissimilar objects, ranking becomes arbitrary regardless of the downstream matching algorithm. To evaluate whether visual embeddings can support enterprise UI search, we analyzed the discriminative power of embeddings through their pairwise cosine similarity distributions.
\textbf{Why Cosine Similarity?} Cosine similarity is the standard metric for high-dimensional embedding spaces because it measures semantic similarity independent of magnitude, focusing on the directional relationship between vectors. In neural embedding spaces, vectors pointing in similar directions represent semantically similar concepts, making cosine similarity the natural choice for evaluating how well embeddings capture meaningful distinctions.
\textbf{Experimental Setup.} We compared six state-of-the-art vision embedding models against our structural approach: three CLIP variants~\cite{radford2021clip} (base, ViT-Base, and base-patch32), two SigLIP models~\cite{SigLip} (large and so400m), and DINOv2-large~\cite{dinov2}. For each model, we computed pairwise cosine similarities across all 20,000 UI screenshots in our dataset. Figure~\ref{fig:embedding_comparison} visualizes the resulting similarity distributions.
\subsubsection{The Representation Collapse Problem}

Our analysis reveals a critical failure mode across all evaluated vision-language models: \textit{representation collapse}, wherein models assign nearly identical embeddings to visually distinct UI screenshots. This phenomenon manifests as extremely narrow similarity distributions concentrated near cosine similarity of 1.0, producing characteristic sharp peaks in density plots, the visual signature of embeddings that have lost discriminative power.

The practical implications of this collapse are severe. When a retrieval system returns cosine similarities of 0.95, 0.96, and 0.97 for three different UI screens, these marginal differences represent measurement noise rather than meaningful similarity rankings. The model effectively cannot distinguish between highly relevant results (e.g., another login screen) and irrelevant ones (e.g., a settings page), degrading the search system to near-random ranking and defeating the fundamental purpose of similarity-based retrieval.

\textbf{Quantitative Analysis.} The severity of representation collapse is evident across all tested models: CLIP-base ($\mu=0.94$, $\sigma=0.041$), CLIP-ViT-Base ($\mu=0.96$, $\sigma=0.032$), DINOv2-large ($\mu=0.89$, $\sigma=0.067$), and SigLIP variants ($\mu=0.87$--$0.91$, $\sigma=0.051$--$0.063$). Standard deviations below 0.07 indicate that 95\% of all pairwise similarities fall within a mere 0.14-point range. Given that meaningful similarity rankings require discriminative power across the full $[0,1]$ interval, such narrow clustering fundamentally eliminates ranking capability.
\subsubsection{Why Vision-Language Models Fail}
This systematic failure across multiple architectures and training regimes is not accidental, it reflects a fundamental mismatch between the models' training objectives and our task requirements.
\textbf{Semantic Generalization vs. Instance Discrimination.} Vision-language models like CLIP are explicitly trained to learn \textit{semantic category representations} that generalize across visual variations. The training objective encourages the model to map "all cats" to similar embeddings regardless of color, pose, or background. This generalization is precisely what makes CLIP excel at zero-shot classification, the model learns to ignore instance-specific details in favor of category-level features.
However, enterprise UI search requires the opposite capability: \textit{instance discrimination}. We need to distinguish between two login screens that are semantically identical (both are "login screens") but structurally different (different button layouts, field arrangements, or navigation patterns). CLIP's training objective actively works against learning these fine-grained distinctions.
\textbf{Natural vs. Synthetic Visual Statistics.} Pre-trained vision models learn feature representations optimized for natural images, photographs of objects, people, and scenes with complex textures, lighting variations, and occlusions. UI screenshots present fundamentally different visual statistics: flat colors, sharp geometric boundaries, text-dominated regions, and hierarchical spatial layouts. The visual features that discriminate between natural images (edge orientations, color gradients, texture patterns) provide minimal signal for discriminating between UI designs.
\textbf{Global vs. Compositional Structure.} Modern vision transformers process images through global self-attention mechanisms that aggregate information across the entire image to form a single embedding. This global pooling works well when the "gist" of an image (e.g., "beach scene" vs. "mountain scene") determines its category. UI screenshots, however, are defined by their \textit{compositional structure}, the spatial arrangement and relationships between components. Two interfaces with identical global statistics (same colors, same amount of text, same basic layout) can serve entirely different functions based on subtle differences in element positioning and hierarchy. Global pooling destroys precisely this compositional information.
\subsubsection{Why Structural Embeddings Succeed}
In stark contrast, our structural embeddings exhibit a broad, roughly uniform distribution spanning cosine similarities from 0.0 to 0.4 (mean 0.18, std dev 0.11). This 2.7× higher standard deviation compared to the best vision model provides meaningful separation across the full similarity spectrum.
\textbf{Task-Aligned Feature Learning.} Our approach learns features specifically optimized for UI discrimination through supervised contrastive learning on UI-specific similarity labels. Rather than forcing the model to generalize across semantic categories, we explicitly train it to discriminate between structurally different interfaces while maintaining similarity for structurally related designs. The training signal directly addresses the task requirements.
\textbf{Compositional Awareness.} By encoding UI screenshots as sets of spatial relationships between detected elements, our structural representation preserves the compositional information that defines interface similarity. Two login screens with similar button arrangements map to similar embeddings even if they differ in colors or text content, while structurally different layouts produce distinct embeddings regardless of surface-level visual similarity.
\textbf{Meaningful Similarity Gradients.} The broad distribution enables fine-grained ranking. Cosine similarities of 0.35, 0.25, and 0.15 reflect genuine differences in structural similarity, progressively less similar layouts, interaction patterns, or element arrangements. Search results can be meaningfully ranked from most to least relevant, with top results substantially more similar to the query than lower-ranked alternatives.
\textbf{Distributional Coverage.} The roughly uniform distribution across [0.0, 0.4] indicates that the embedding space effectively utilizes its representational capacity. Rather than clustering all points in a narrow region, embeddings spread across the space to maintain maximum separability. This full utilization of the embedding space is a hallmark of well-learned discriminative representations.
\subsection{Performance and Scalability}
    
    From 950 measured operations, overall average latency was 47.30ms (median 47.48ms), with minimum 0.053ms and maximum 262.51ms. Breakdown by query type:
    \begin{itemize}
        \item Metadata and structural: $\sim$0.3ms
        \item Semantic: 49.16ms
        \item Complex multimodal: 19.75-158.17ms
    \end{itemize}
    
    The empirical cumulative distribution shows P50 at 47.5ms, P90 at 108.9ms, P95 at 123.7ms, and P99 at 232.0ms. Most queries complete well under 125ms, enabling interactive user experiences.
    
    Resource utilization remains efficient: CPU usage averaged 6.9\% (peaks 36.2\%), GPU usage averaged 19.3\% (peaks 50\%), and memory averaged 0.6MB per query with peak at 342.9MB. GPU acceleration reduced semantic search latency by approximately 60\%.
    
    
    \begin{table}[t]
    \centering
    \caption{Scalability validation across collection sizes.}
    \label{tab:scalability}
    \begin{tabular}{@{}lcccc@{}}
    \toprule
    Collection & Index & Memory & Query & Cache Hit \\
    Size & Time (s) & (GB) & Time (ms) & Rate \\
    \midrule
    1,000 & $<$0.1 & 0.01 & 0.1 & 0.92 \\
    5,000 & $<$0.1 & 0.05 & 0.2 & 0.94 \\
    20,000 & 0.3 & 0.20 & 0.3 & 0.96 \\
    \bottomrule
    \end{tabular}
    \end{table}

    

\section{Threats to Validity}
We discuss internal, external, construct, and reliability threats, stating assumptions and mitigations to calibrate the scope of our claims.

\textbf{Internal Validity.} The primary threat to internal validity is the dependency on the accuracy of the object detection model . Detection errors (estimated 5-8\% based on YOLO's typical performance on UI datasets) propagate through graph construction, potentially introducing noise in structural embeddings. We mitigate this through robust graph construction with IoU-based edge creation that can handle minor localization errors. However, missed elements or misclassifications directly impact graph topology and subsequent similarity computations.

The multi-level similarity function uses manually tuned weights ($\alpha=0.6, \beta=0.3, \gamma=0.1$) which may not generalize optimally across all UI types. While these weights were validated on our financial UI dataset, different UI domains might benefit from alternative weightings. The contrastive learning threshold ($\theta_{\text{pos}}$) also influences positive pair selection, and suboptimal values could degrade embedding quality.

\textbf{External Validity.} Our evaluation focuses exclusively on a single enterprise financial UI, which represents a specific subset of enterprise software with consistent design patterns and domain-specific elements. Generalization to other domains (e.g., mobile apps, consumer web applications, or design tools) remains unvalidated.

The dataset of 20,000+ screens, while substantial, represents a single organization's design system. Cross-organizational variation in terminology, styling conventions, and structural patterns may affect performance. Our approach assumes English text; multilingual UIs with mixed scripts would require OCR adaptations.

\textbf{Construct Validity.} Our evaluation relies heavily on automated VLM judges (12 models with 89.2\% acceptance rate), which may not perfectly align with human judgments of UI similarity. While VLMs provide scalable evaluation, they may exhibit systematic biases not present in human assessments. We partially address this through ensemble evaluation across diverse VLM architectures.

\textbf{Reliability.} Query latency measurements were performed on dedicated hardware with controlled load. Production deployments with concurrent users, network latency, and resource contention may exhibit different performance characteristics. The reported P95 latency of 124ms represents best-case infrastructure and may not generalize to resource-constrained environments.

\section{Conclusions}
    
    We presented the first unified multimodal retrieval framework combining object detection-driven metadata filtering, graph-based structural similarity, and semantic embedding search for enterprise UI collections. Our system achieves 0.92 Top-5 accuracy with 47.5ms median latency on 20,396 financial UIs. 
    
    Numerical analysis reveals that our graph-based structural embeddings has a significantly better discriminative power than state-of-the-art vision encoder. This enables fine-grained UI distinction impossible with currently used approaches. This improvement stems from explicitly modeling hierarchical structure and spatial relationships through graph neural networks with multi-head attention and contrastive learning.
    
    The hybrid indexing architecture combining FAISS vector search with specialized metadata indices supports complex multimodal queries through a composable query language, enabling queries like ``checkout screens structurally similar to template.png with 2-5 input fields but no tables.'' Query optimization strategies achieve subsecond latency through cost-based index selection and dynamic similarity fusion.

{
    \small
    \bibliographystyle{plain}
    \bibliography{static/references}
}
\end{document}